# Approximate Computing for Robotic path planning - Experimentation, Case Study and Practical Implications

Hrishav Bakul Barua [1]

## Abstract

Approximate computing is a computation domain which can be used to trade time and energy with quality and therefore is useful in embedded systems. Energy is the prime resource in battery-driven embedded systems, like robots. Approximate computing can be used as a technique to generate approximate version of the control functionalities of a robot, enabling it to ration energy for computation at the cost of degraded quality. Usually, the programmer of the function specifies the extent of degradation that is safe for the overall safety of the system. However, in a collaborative environment, where several sub-systems co-exist and some of the functionality of each of them have been approximated, the safety of the overall system may be compromised. In this paper, we consider multiple identical robots operate in a warehouse, and the path planning function of the robot is approximated. Although the planned paths are safe for individual robots (ı.e. they do not collide with the racks), we show that this leads to a collision among the robots. So, a controlled approximation needs to be carried out in such situations to harness the full power of this new paradigm if it needs to be a mainstream paradigm in future.

Robotics and Autonomous Systems Research Group
Cognitive Robotics & Vision
TCS Research [1]
Kolkata, India

# 1 Introduction

In recent years *in-exact* computing or *approximate* computing has emerged as an alternative technique to address timing and energy issues in embedded systems. Based on earlier works by Palem *et al* [9], this paradigm has attained attention of the researchers around the world. At present a considerable amount of fundamental work in this area is available. A comprehensive study of research works, in both the fields of hardware and software design can be found in [25, 18].

From the perspective of approximate computing, modern systems are over-provisioned for accuracy and therefore they spend more than necessary time and energy to achieve a "correct" output. The underlying philosophy is to sacrifice enough accuracy from computation and gain in terms of time or energy or both. The trade-off is essentially an optimization problem, which attempts to inject as much imperfections as possible, to gain time and/or energy without violating the error tolerance margin set by the programmer. There are a variety of applications which are inherently tolerant to approximation, *eg* computer vision, media processing, machine learning.

Approximate computing has very practical implication in embedded systems, where both time and energy are very costly resources. For example, in robotics, execution time of an application running in it directly determines the response latency of the robot [6]. Again many robotics systems are battery-powered. In such systems, both time and energy required for computation of the application need to be bounded. One can aggressively approximate the application to achieve tighter timing and energy ceilings. Error tolerance level supplied for approximation usually ensures tolerable working condition of a single system.

However, in present day, we see multiple embedded system works in a common space and also interact. Such collaborative execution of tasks by a group of robots is evident in many warehouse floors. If approximate computing methodology is adopted in such a system, without consideration of the approximation strategy adopted by co-working robots, the overall safety and correctness of the system may be compromised.

Let us consider a set of robots working in an warehouse floor. The warehouse corridors are designed to be wide enough to allow two robots to move side-by-side. A centralized controller computes paths for each of the robots. Figure 1(a) show exact computation of the path of the two robots which is collision free. Consider we apply in-exactness in the computation of path and allow, let us say 10% deviation, as error margin in the approximate version. In Figure 1(b) paths deviations negatively interfere with each other where both the robots sway in the same direction such that the relative distance between them remains the same. In this case the paths are safe. However, in some cases the path interference become positive, *ie* the robots sway towards each other as shown in Figure 1(c) and result in a collision.

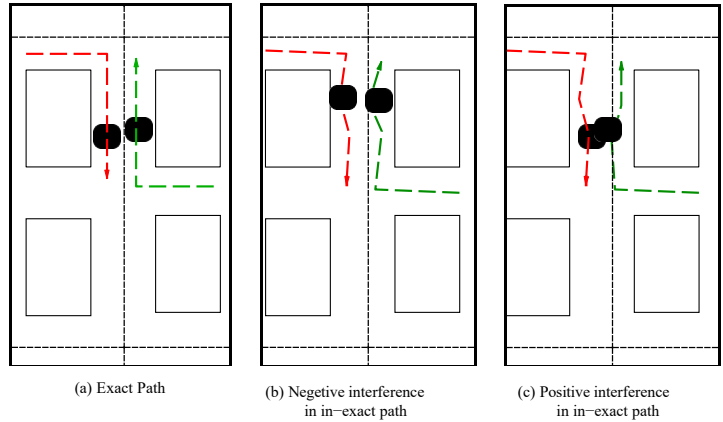

(a) Exact Path    (b) Negetive interference in in−exact path    (c) Positive interference in in−exact path

Fig. 1: Collision in floor due to in-exactness

**Motivation:**

Collaborative systems such as Multi-robot systems are mostly battery powered where individual working sub-systems are assigned with their part of the task. From the computational aspects of such systems, many resource and energy hungry tasks needs to be performed and on a frequent basis. One of such task is *Robot path planning* in a multi-robot collaborative environment containing 2 or more robots in an industry warehouse. Robots are expected to take the optimum path from their start point to the destination without colliding with the obstacles (racks, objects and other robots etc.). Many methods and algorithms are well in place to attain this feat. Energy and execution time of such tasks may be crucial in battery powered systems where tasks have deadlines. To solve these open optimization problems in collaborative system we have chosen Approximate (In-exact) computing as a paradigm and intend to put forward its implications in collaborative environments.

## 2 Literature review of related works

*Approximate* or *In-exact* computing is around from the 90's and really gained importance by the beginning of the current decade. A systematic survey on this subject can be found in [18, 5] which describes methods, techniques and tools for application of approximation. The basic idea of approximate computing has been nicely explained in [13, 19]. Both hardwares and softwares can be approximated with different methods and approaches shown in [25].

Some of the available tool chains and compilers for approximate generation of exact code have been reviewed. ACCEPT [26] is a C compiler inspired by LLVM [16, 11]. ACCEPT uses multiple approximation techniques such as *loop* *perforation*, *precision scaling* and *memoization* [18] which are in-built in the AC-CEPT's system. It uses annotations to identify approximative portions of C/C++ codes. *Approx* and *Endorse* keywords are used in codes to teach the compiler which portions of the code it should try to approximate. It creates an interactive loop between the user and the compiler where the compiler provides feedback on various approximate profiles of the code as per the annotations given by the user. The user can continue this process till the compiler finds some good approximate versions of the code with a balance between energy efficiency and latency in one hand and quality and accuracy of output on the other. Some other tools with similar functionalities but for Java based codes are EnerJ [27] and FlexJava [20]. React [32] is an extension of ACCEPT which provides efficient approximation support for fine grain as well as coarse grain level in codes.

This part provides a brief compilation of the widely used techniques and methods in Approximate Computing as per the available literature. *Loop perforation* [29, 26] is an approach used in many application such as video/audio processing, speech processing, image processing, machine learning and big data mining. The general idea of loop perforation is dropping some loop iterations from the original execution of loop with the intention of getting speedup of the entire execution bearing some tolerable loss of output quality. For example, in case of image processing if some pixels are periodically missed while processing through loops then it will degrade the output image quality but still acceptable as the overall meaning of the image sustain. Some of the loop perforation related researches are [10, 17, 2, 4, 28]. One of the current technique for approximation is *Memoization* [18]. The concept of this approach is using already computed results in an application for similar functions or inputs. Here, approximation can be expected in such a way that for near similar inputs of a program the previously computed results will be given to the user without re computation of the same. This can be explored for various levels of granularity in a big program and similarity ratio between previous and new input can be considered. Many *Spatial* and *Temporal* memoization techniques have been discussed in [1, 22, 24, 23, 33]. Profiling of a code, in the perspective of approximation, can yield multiple versions of the exact code in approximate form. These versions will result in different energy-quality trade-offs. Some of the works in this direction are [3, 4, 12, 31]. Some of the works are focused on skipping tasks and memory accesses to reduce overhead of computation resulting in energy saving but also leading to QOR loss. The user is expected to identify the tasks which have least impact on the overall output of the application as candidate. Some of the notable works in this context can be found in [24, 12, 7, 30, 8]. Another method for approximation is *Precision scaling*. In this technique, the operands of an operation are altered bit-width wise to reduce computing and storage load. Generally lower order bits are dropped or altered. The operands for input data and intermediate data needs to be identify as candidate depending on their criticality and impact on overall success of the application. The works concentrated on this area are [33, 30].

Annotating the code with special keywords has facilitate the users to have control over the approximation activity performed by the compilers and tool-chains. Probable candidate for approximation needs to be identified by a user and annotate with the keywords provided by the approximate compiler. Bit level approximation can be performed for image processing and video data. Finding probable approximation and approximable portions in code can be accomplished by error injection also. Some of the pioneering works in this category can be found in [26, 27, 20, 3, 21]

## 3 Approximation, Experimental Setup and Results

Our experimental setup is based on warehouse operation system where multiple robots serves as workers. The basic task assigned to these robots are to pick a set of objects from various locations in the warehouse, and carry them to their pre-defined locations or *dock*. The double problem of *task assignment* and *path planning* in this context are related. The problem of *task assignment* deals with partitioning a set of task and assign each partition to one robot such that some objective is minimized. One of the most interesting optimization criterion is the overall completion time of the tasks. Naturally, in the context of this optimization criterion, planning the path of each of the robots emerges as a sub-problem to the task allocation problem. It has been shown that this problem, with most of their non-trivial variations, are in NP.

In our experiments, we simplify the task assignment problem as a single task to single robot assignment, which can be efficiently solved using *Hungarian Method* [15]. We employ *A\** [14] as the path planning algorithm, which is also the most popular method for planning path for a robot. In our experiment, we apply approximation on the path planning algorithm. Also we follow the usual conditions for application of approximation *ie* the quality degradation of computed path is bound within very small range. And also we ensure that the path does not force the robot to collide with static obstacles in the warehouse due to approximation. We also assume that the robots start their individual journey at the same time and can travel one grid cell at one time unit. The robots are considered to be homogeneous.

### 3.1 Background

Some of the commonly used methods for generating approximate version of a software are *loop perforation*, *precision scaling* and *memoization*. We use loop perforation in our experiment on *A\**. Loop perforation basically refers to dropping off some loop iterations either randomly or following a pattern. This is categorized into three types [29] - Modulo perforation, Truncation perforation and Random perforation and they have different perforation rate. The quality of the output depends upon the perforation rate - higher the perforation rate higher the gain in time but also higher the quality degradation.

*Modulo perforation* is a perforation model where loops are skipped in a pattern. In the following case $n - 1$ loops are missed in each iteration, perforation rate is represented as $(n-1)/n$.

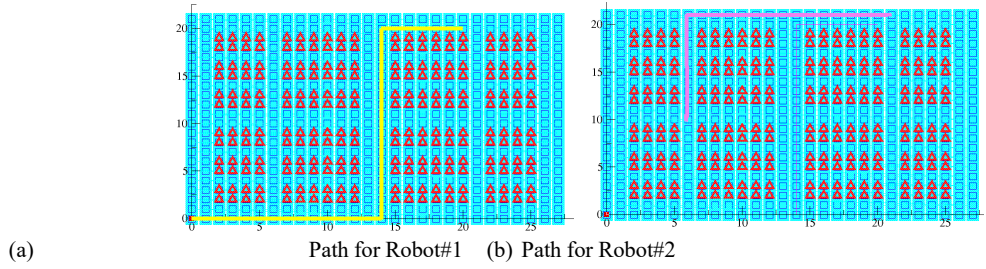

(a)                             Path for Robot#1          (b)  Path for Robot#2

Fig. 2: Exact planning of paths

```
for(i=0; i<m; i+=n) { ... }
```

Whereas, in the following perforation, loops are missed in the multiples of *n* iterations *ie* 1 in every *n* loops are missed, perforation rate is $1/n$.

```
for(i=0; i<m; i++) { if(i%n==0) continue; ... }
```

The exact loop looks like:

```
for(i=0; i<m; i+=1) { ... }
```

*Truncation perforation* is a method of loop skipping where a series of continuous loop iterations are missed in the beginning or at the end depending upon the application and criticality of the loop.

*Random perforation* is the technique where loop iterations are dropped without any pattern unlike the other two methods mentioned above.

To ensure the quality of service, metric considered for our path planning case is dependent on the deviation of the approximate path length from the exact path length. The equation 1 below gives the average percentage increase in path length for many arbitrary runs of $A*$ for different source and destinations of a robot.

$$E_p = \frac{1}{n} \left( \frac{A(P_L) - O(P_L)}{O(P_L)} \right) \times 100 \tag{1}$$

A($P_L$) is the approximate path length for a particular case. O($P_L$) is the original path length for a case and n is the total number of arbitrary cases.

## 3.2 Experimental Setup

Our experiment is carried out in two steps. We use module perforation technique for creation of approximate version. We experimented with various portions of the code and checked with different functions and loops. We have chosen the most suitable loop for perforation which we determine by thoroughly experimenting with each and every loop. The first step of the experiment is to determine the perforation ratio which maximizes gain in time and keeps the quality degradation within a specified range. Once we determine the perforation ratio and thereby the corresponding approximate version of the $A*$ algorithm, it is used in the second phase of the experiment. In the second phase of the experiment we create paths of multiple robots in a warehouse using the approximate version of $A*$ and show that in some cases two robots collides with each other, although individually their paths are clean *ie* they do not collide with static objects in the warehouse.

### 3.2.1 Determination of suitable Perforation rate

We have conducted a rigorous experimentation to establish the perforation rate optimum for our cases. We have collected data for perforation rates ranging from *0.2* to *0.88* and checked the speedup and error in paths for many cases.

Table 1: Perforation rate and Speedup data collected from experimenting $A*$

| Perforation rate | Achieved average speedup | Percentage of cases with increase in path length | Percentage of cases where the algorithm fails |
|---|---|---|---|
| 0.2 | 1.85X | 0 | 0 |
| 0.25 | 1.97X | 0 | 0 |
| 0.33 | 2.04X | 0 | 0 |
| 0.5 | 2.404X | 5 | 0 |
| 0.6 | 2.748X | 5 | 0 |
| 0.75 | 2.942X | 10 | 0 |
| 0.8 | 3.155X | 15 | 0 |
| 0.83 | 3.1X | 10 | 5 |
| 0.85 | 3.36X | 20 | 15 |
| 0.88 | 3.4X | 20 | 5 |

The data in Table 1 is generated by considering various random start and end point for a robot in a Warehouse scenario shown in Figure 2 where each rectangular grid cell is a *2-D* coordinate position for a robot. From the generated data we can choose a perforation rate for the path planning activity of robots considering the speedup and path error. The chosen rates are *0.6*, *0.75* and *0.8*. The data collected for these rates also predicts that maximum increase in path length for these rates (in a few cases) are not more than 10%. Figure 3 (a) shows the speedup distribution over the approximate execution of $A*$ for arbitrary paths and Figure 3 (b) shows the bar graph of speedup against loop perforation rates for arbitrary paths. Figure 4 shows the lengths of arbitrarily chosen paths and the speedups achieved for various Perfo-ration rates (denoted as PR in the graph). Figure 5 (a) shows the trend of achieved speedup VS perforation rate calculated for a number of arbitrary cases. Figure 5 (b) is an analysis of the growing error cases in the system with increasing perforation rate.

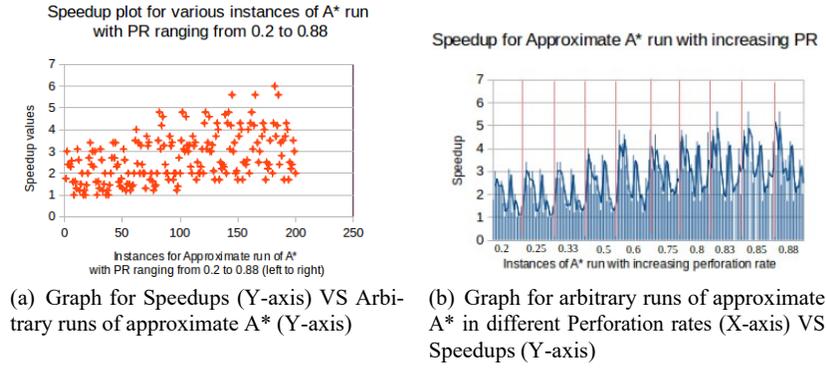

(a) Graph for Speedups (Y-axis) VS Arbitrary runs of approximate A* (Y-axis)

(b) Graph for arbitrary runs of approximate A* in different Perforation rates (X-axis) VS Speedups (Y-axis)

Fig. 3: Experimental Results with Perforation rates

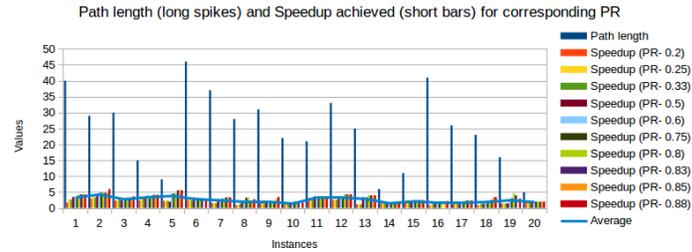

Fig. 4: Experimental Results with arbitrarily chosen paths for varied Perforation rates

### 3.2.2 Implications on Approximation of $A*$ with Loop perforation

Here we have made an attempt to use the selected perforation rates in two cases. One is the typical Warehouse scenario as shown in Figure 2 where two robots will operate individually and follow their pre-computed paths. The other one is a simple room scenario where we have a table in the middle and two/four robots will go towards the table from corners to collaborate. This is shown in Figure 6.

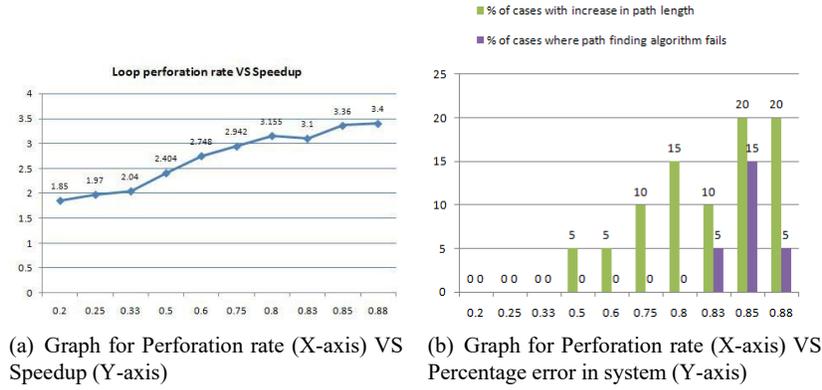

(a) Graph for Perforation rate (X-axis) VS Speedup (Y-axis)

(b) Graph for Perforation rate (X-axis) VS Percentage error in system (Y-axis)

Fig. 5: Experimental Results with Perforation rates

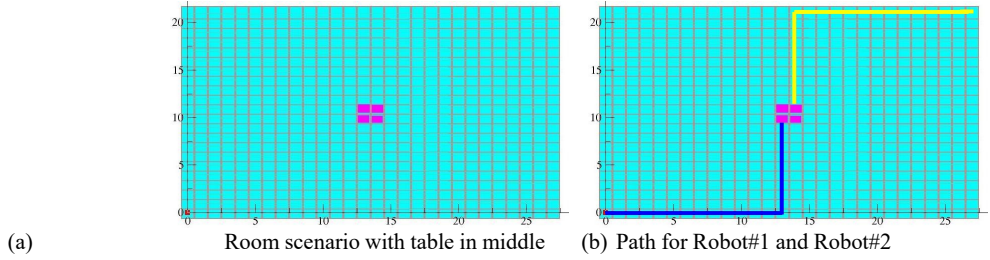

(a) Room scenario with table in middle

(b) Path for Robot#1 and Robot#2

Fig. 6: Exact paths in Room scenario

For the chosen rates, we conduct experiment on warehouse scenario and find that 10-20% of the cases shows collision between the robots. Table 2 shows this data.

Table 2: Perforation rate and Speedup data collected from experimenting $A*$ in Warehouse scenario

| Perforation rate | Achieved average speedup | Percentage of Collision cases encountered |
|---|---|---|
| 0.6 | 2.748X | 5 |
| 0.75 | 2.942X | 20 |
| 0.8 | 3.155X | 20 |

For perforation rate *0.75*, we have shown the *Exact* and *In-exact* cases for 2 robots where the latter case leads to collision.

### *Exact path planning for Warehouse Scenario:*

We consider two robots, one starts from *grid(5,4)* and destined for *grid(21,19)* and the other starts from *grid(2,7)* and destined for *grid(14,19)*. This case is shown in Figure 7.

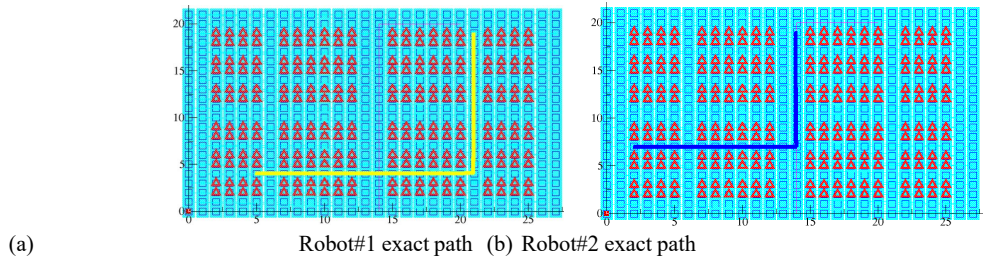

(a)       Robot#1 exact path    (b) Robot#2 exact path

Fig. 7: Exact paths in Warehouse scenario

The robots have completed the paths without any collision as clearly seen in Figure 9 (a).

### *In-exactness in path planning for Warehouse Scenario:*

We now show the case after introducing *in-exactness* in path planning activity using loop perforation. The case is shown in Figure 8.

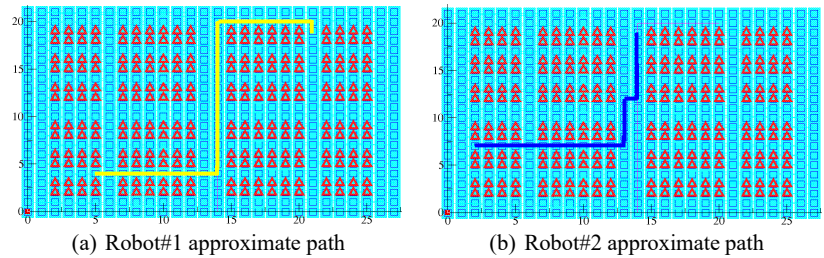

(a) Robot#1 approximate path      (b) Robot#2 approximate path

Fig. 8: Approximate paths in Warehouse scenario

We see that the paths are near optimum where the path of robot#1 has increased by 2 hops and the length of path for robot#2 remains the same although they have taken different paths.

We also see that these *in-exact* paths collide at a point (*t=17*). This is shown in Figure 9 (b)

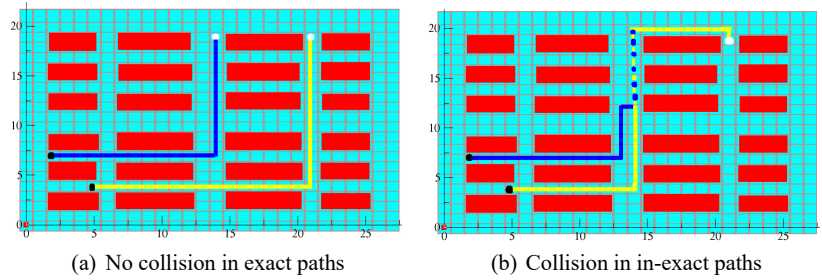

(a) No collision in exact paths       (b) Collision in in-exact paths

Fig. 9: Collision VS No-collision in Warehouse scenario

A detailed figure with timestamps is shown in Figure 10 (a). Here, we have shown that both the robots starts at *t=0*. They travel in their planned paths (as per the path planning algorithm) and finally collide at *t=17*.

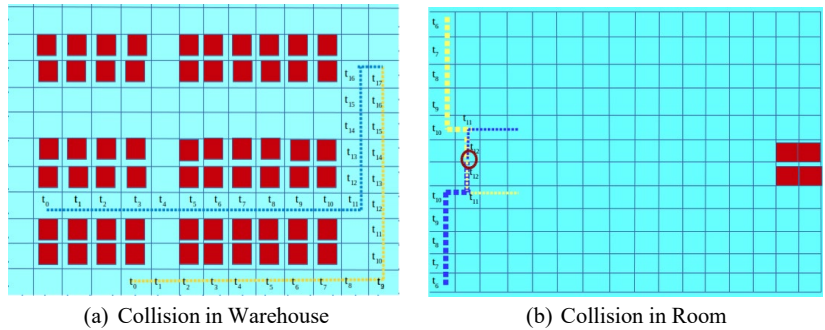

(a) Collision in Warehouse       (b) Collision in Room

Fig. 10: Collision with timestamps

### *Exact path planning for Room Scenario:*

Here, we consider two robots coming to the center of the room in Figure 6. The room has a table in the middle. The robots will perform a handshaking activity or exchange objects and proceed to their destinations. We consider the same perforation rate *0.75*. The first robot starts at *grid(0,0)* and reaches the table at *grid(12,10)*, after handshaking it goes to its final destination *grid(27,0)*. Similarly, second one starts at *grid(0,21)* meets robot#1 near the table at *grid(12,11)* and finally heads for *grid(27,21)*. These cases are shown in Figure 11. The Robot#1 takes the blue path and yellow for Robot#2. The gray dots are the intermediate destination for the robots and the black ones are the final destination after handshaking.

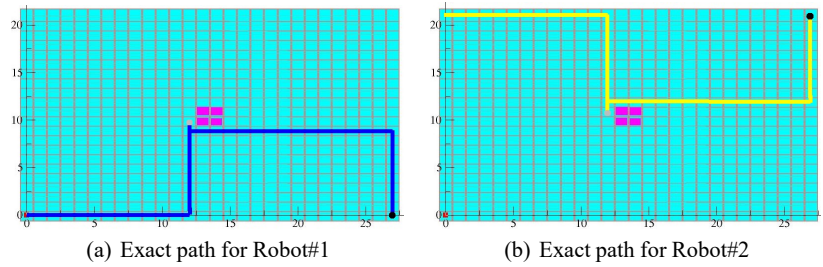

(a) Exact path for Robot#1           (b) Exact path for Robot#2

Fig. 11: Exact paths in Room scenario

The robots have reached their intermediate and final destinations without collision as shown in Figure 13 (a).

### In-exactness in path planning for Room Scenario:

We now intend to show the *in-exact* paths for the robots in room scenario. This case is shown in Figure 12.

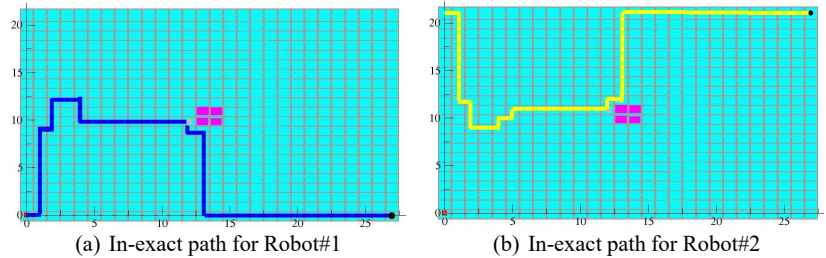

(a) In-exact path for Robot#1         (b) In-exact path for Robot#2

Fig. 12: In-exact paths in Room scenario

The approximate paths differ from the exact one by a few hops. We have considered the same loop perforation rate as in Warehouse scenario. We can see in Figure 13 (b) that the approximate paths collide at a point (*t=13*).

A detailed view is shown in Figure 10 (b). Both the robots start at *t=0* (starting points not shown; shown from *t=6*), and head-on collides at *t=13*.

In some of the cases for Warehouse and Room scenario the path lengths have increased and collision cases are encountered. Figure 14 shows the behavior of the path planning system (*ie* path length deviation and collisions) on application of approximation using loop perforation.

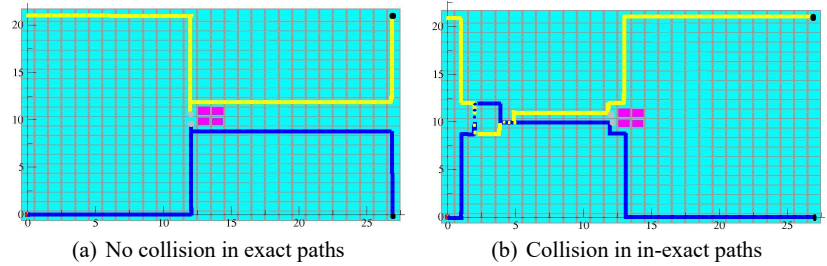

(a) No collision in exact paths          (b) Collision in in-exact paths

Fig. 13: Collision VS No-collision in Room scenario

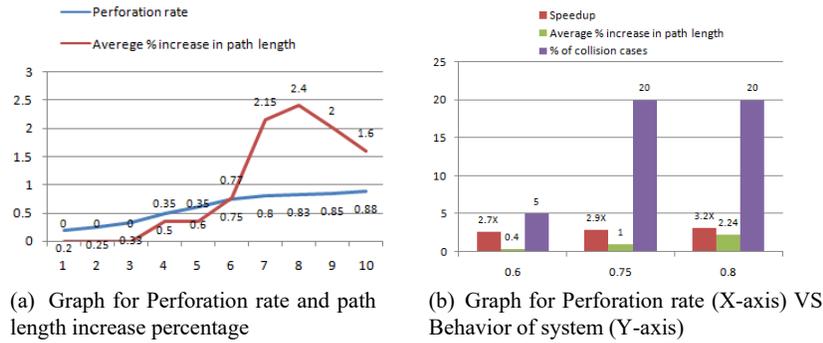

(a) Graph for Perforation rate and path length increase percentage

(b) Graph for Perforation rate (X-axis) VS Behavior of system (Y-axis)

Fig. 14: Experimental Results on System behavior

## 4  Conclusion and Future scope of work

Collaborative systems (in robotics) have gained much importance in the recent times for industry warehouse maintenance and other tasks which needs collaboration between robots and humans. But these systems are generally battery powered. The tasks performed by such systems needs high computational resources. If we can compromise the overall safety and quality of the system then we can achieve good speedup in computations and energy savings. This paper introduces *Approximate* computing (*In-exact*) in the *path planning* activity of a simple multi-robot system. We have used *loop perforation* technique in approximating *A\** and experimented rigorously before choosing the best *perforation* rates considering the speedup gain and path deviations. This paper puts forward the dangerous implications of approximate computing in path planning as some of the cases lead to collision between two robots although the robots don't collide with the warehouse racks or other static obstacles. Finally, the paper states that if the quality of output and safety of the entire system is compromised, we can achieve better turnaround time for path planning computation and save a considerable amount of computational energy. Future scope of research in this area will need focus on integrating existing tool-chains and ap-proximate compilers in the compute layer of such collaborative systems and achieve approximate results dynamically (degraded to an extend safe for the overall working of the system) for various functions apart from robot path planning.